# ChestNet: A Deep Neural Network for Classification of Thoracic Diseases on Chest Radiography

HongyuWang, Yong Xia*

1Shaanxi Key Lab of Speech & Image Information Processing (SAIIP), School of Computer Science and Engineering, Northwestern Polytechnical University, Xi'an 710072, China

Corresponding Author's Email: yxia@nwpu.edu.cn

**Abstract.** Computer-aided techniques may lead to more accurate and more accessible diagnosis of thorax diseases on chest radiography. Despite the success of deep learning-based solutions, this task remains a major challenge in smart healthcare, since it is intrinsically a weakly supervised learning problem. In this paper, we incorporate the attention mechanism into a deep convolutional neural network, and thus propose the ChestNet model to address effective diagnosis of thorax diseases on chest radiography. This model consists of two branches: a classification branch serves as a uniform feature extraction-classification network to free users from troublesome handcrafted feature extraction, and an attention branch exploits the correlation between class labels and the locations of pathological abnormalities and allows the model to concentrate adaptively on the pathologically abnormal regions. We evaluated our model against three state-of-the-art deep learning models on the Chest X-ray 14 dataset using the official patient-wise split. The results indicate that our model outperforms other methods, which use no extra training data, in diagnosing 14 thorax diseases on chest radiography.

## 1    Introduction

Thorax diseases is a major health thread on this planet. The pneumonia alone affects approximately 450 million people (i.e. 7% of the world population) and results in about 4 million deaths per year [1]. Due to its low-cost and easy-access nature, chest radiography, colloquially called chest X-ray (CXR), is one of the most common types of radiology examinations for the diagnosis of thorax diseases. The enormous number of chest radiographs produced globally are currently analyzed almost entirely through visual inspection on a slice-by-slice basis. This requires a high degree of skill and concentration, and is time-consuming, expensive, prone to operator bias, and unable to exploit the invaluable informatics contained in such large-scale data [2]. Moreover, due to the complexity of chest radiographs, it is challenging even for radiologists to discriminate thorax diseases on them, resulting in the shortage of expert radiologists who are competent to read chest radiographs in many countries. Therefore, it is of significance to develop automated algorithms for the computer-aided diagnosis of thorax diseases on chest radiography.

In recent years, deep learning techniques have achieved profound breakthroughs in many computer vision applications, including the classification of natural and medical images [3-5]. This success has prompted many investigators to employ deep convolutional neural networks (DCNNs) for the diagnosis of thorax diseases on chest radiography. Wang et al. [6] proposed a unified weakly-supervised multi-label classification framework by considering various multi-label DCNN losses and different pooling strategies. Since there may be multiple pathological patterns on each chest radiograph, Yao et al. [7] developed an approach that further exploits the statistical label dependencies, and thus achieved improved performance. Similarly, Kumar et al. [8] employed multi-label learning techniques and investigated the potential dependencies among the labels. Rajpurkar et al. [9] designed an deep learning model called CheXNet, and used dense connections [10] and batch normalization [11] to make the optimization of such model tractable. According to the results they reported, CheXNet is able to detect pneumonia at a level matching or exceeding radiologists.

Intuitively, computer-aided diagnosis of thorax diseases consists of two successive steps: the detection of pathological abnormalities and classification of them. Automated abnormality detection on chest radiographs is also a challenging task, due to the complexity and diversity of thorax diseases and the limited quality of chest radiographs. Manually marking the counters of abnormal regions on a chest radiograph requires even more work than labelling it. Hence, few public chest radiography data are equipped with the mask of abnormalities [6], leading this computer-aided diagnosis task to a weakly supervised problem (i.e. giving only the names of abnormalities existed in each radiograph but no locations of them). Recently, attention-based models have demonstrated their proven ability to learn the localization and recognition of multiple objects despite being given only class labels during training [12]. Therefore, we suggest using the attention mechanism to address this weakly supervised learning challenge.

In this paper, we propose a deep neural network model called ChestNet for the computer-aided diagnosis of thorax diseases on chest radiography. This model consists of two branches: a classification branch serves as a uniform feature extraction-classification network to free users from troublesome handcrafted feature extraction, and an attention branch exploits the correlation between class labels and the locations of pathological abnormalities via analyzing the feature maps learned by the classification branch. Feeding each chest radiograph to the model, the diagnosis is obtained by averaging and binarizing the outputs of both branches. The proposed ChestNet model was evaluated against three deep learning models on the Chest X-ray 14 dataset using the official patient-wise split, and achieved the state-of-the-art performance.

## 2  Dataset

As a hospital-scale chest radiography dataset, ChestX-ray14 [6] consists of 112,120 frontal-view X-ray images acquired from 30,805 unique subjects with 14 disease image labels including atelectasis, cardiomegaly, effusion, infiltration, mass, nodule, pneumonia, pneumothorax, consolidation, edema, emphysema, fibrosis, pleural thickening and hernia. These images were originally saved in the PNG format and were then rescaled to the size of $1024 \times 1024$. Each image has one or multiple labels, which were mined from the associated radiological reports using natural language processing. The dataset was officially split into a training subset (80% subjects) and a testing subset (20% subjects) on the patient level. All studies from the same patient appear only in either the training or testing subset.

## 3  Method

The proposed ChestNet model (see Fig. 1) consists of a classification branch and an attention branch. We now delve into each branch.

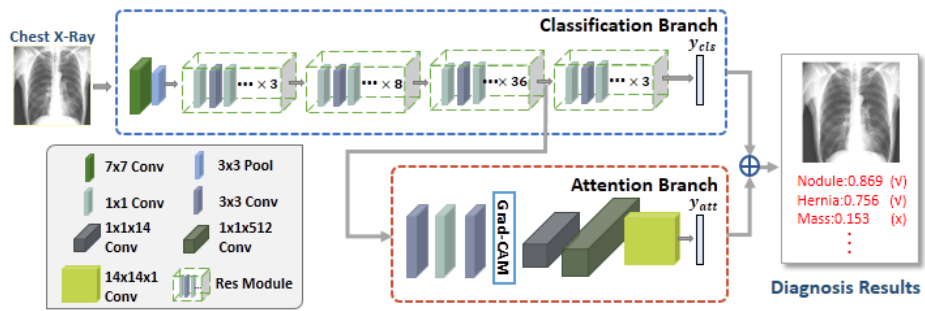

**Fig. 1.** Diagram of the proposed ChestNet model

### 3.1  Classification Branch for Label Prediction

The classification branch is the ResNet-152 model [3], which has been pre-trained on the ImageNet dataset. ResNet-152 consists of 152 learnable layers, including consequently a convolutional layer with the kernel size of 7x7, a 3x3 max pooling layer, and four residual modules, which have 3, 8, 36, and 3 triple-layer residual blocks, respectively. To adapt ResNet-152 to our problem, we removed the softmax layer and replaced the last fully connected layer with a layer of 14 neurons, each using the sigmoid activation. We defined the outputs of these 14 neurons as a label prediction vector $\mathbf{y}_{cls}$.

### 3.2    Attention Branch for Abnormality Detection

The attention branch is used to exploit the correlation between class labels and the regions of pathological abnormalities via analyzing the learned feature maps. For this study, we chose the output of the penultimate residual module in ResNet-152 as the input of the attention branch.

The attention branch is composed of 6 convolutional layers. The first 3 convolutional layers use 1x1, 3x3, and 1x1 kernels, each being followed by a ReLU activation function. Based on the output of the third convolutional layer, denoted by $\widetilde{\mathbf{A}}$, we employed the gradient-weighted class activation mapping (Grad-CAM) [13] to estimate the class discriminative localization map $\overline{\mathbf{A}}$ for each class c, shown as follows

$$\overline{\mathbf{A}}_c = \text{ReLU}(\textstyle\sum_k \alpha_{ck} \widetilde{\mathbf{A}}_k), \tag{1}$$

where $\alpha_{ck}$ can be computed by using the gradient propagation [13]. Then, we normalize each element of $\overline{\mathbf{A}}_c$ as follows

$$\text{a}_{cij} = \exp(\overline{\text{a}}_{cij}) / \textstyle\sum_{i,j} \exp(\overline{\text{a}}_{cij}), \tag{2}$$

And we used the normalized class discriminative localization map $\mathbf{A}$ as the input of the fourth convolutional layer. Following the design in [14], the last 3 convolutional layers contain 14 1x1 kernels, 512 1x1 kernels and a 14x14 kernel, respectively. The 4th and 5th layers use ReLU as the nonlinearity operator, and the last layer uses the sigmoid activation. We defined the attention branch's output as the label confidence vector $\mathbf{y}_{\text{att}}$.

### 3.3    Training and Testing

To train the proposed ChestNet model, we resized the chest radiographs to the size of $224 \times 224$ such that they can be fed into the pre-trained ResNet-152, but did not perform data augmentation. We adopted the official patient-wise split of the ChestX-ray14 dataset [6] and used the images from 10% subjects in the training set to form a validation set. We first trained the classification branch, then trained the attention branch with all parameters in the classification branch being fixed, and finally fine-tuned the proposed ChestNet model in an end-to-end manner. Each training process aims to minimize the following cross-entropy loss

$$\text{L}(\mathbf{y}, \mathbf{y}^{(P)}) = \textstyle\sum_{c=1}^{14} \text{y}_c \log(\text{y}_c^{(P)}) + ((1 - \text{y}_c) \log(1 - \text{y}_c^{(P)}), \tag{3}$$

where $\mathbf{y}$ is a 14-dimensional ground-truth vector in which each element is binary and a "1" represents the existence of the corresponding thorax disease, and $\mathbf{y}^{(P)}$ is the predicted label vector, which represents the label prediction vector $\mathbf{y}_{\text{cls}}$ when training the classification branch, the label confidence vector $\mathbf{y}_{\text{att}}$ when training the attention branch, and the average of $\mathbf{y}_{\text{cls}}$ and $\mathbf{y}_{\text{att}}$ when training the ChestNet model. We adopted the mini-batch stochastic gradient descent algorithm with a batch size of 24 and momentum to 0.9 as the optimizer, and set the gamma value to 0.1, the learning rate to 0.001, the decay of weight to 0.0005, and the maximum iteration number to 50000.

After feeding a test image into the trained ChestNet model, a 14-dimensional prediction vector $\mathbf{y}^{(P)}$ is obtained. If any elements of $\mathbf{y}^{(P)}$ are greater than the threshold 0.5, the diagnosis of the corresponding thorax diseases is positive.

## 4 Results

Fig. 2 shows two example test images and the diagnosis made by the proposed ChestNet model. The first case was diagnosed with the atelectasis and effusion, and the second case was diagnosed with the effusion and infiltration. The obtained receiver operating characteristic (ROC) curve of diagnosing each thorax disease was shown in Fig. 3.

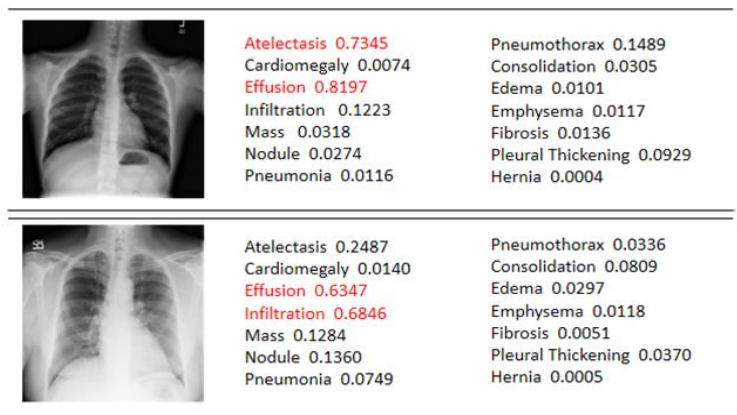

**Fig. 2.** Two test images and their diagnoses made by the proposed ChestNet model

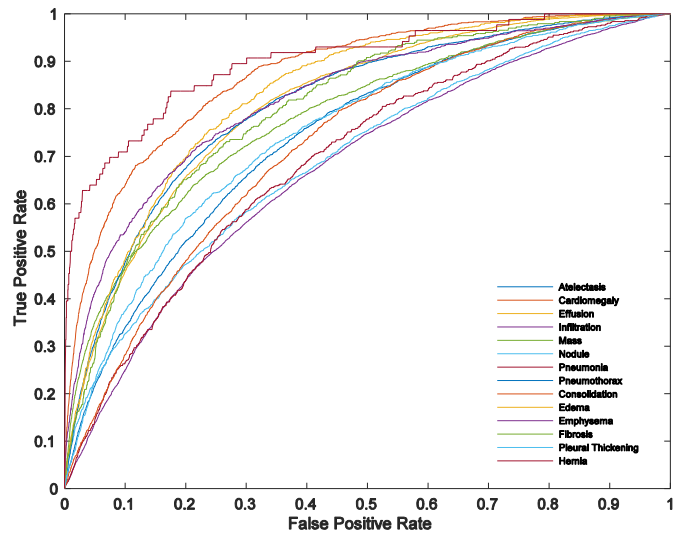

**Fig. 3.** ROC curves of ChestNet for each of 14 thorax diseases

Next, we compared our ChestNet model to three state-of-the-art deep learning methods. The per-class area under the ROC curve (AUC) obtained by applying these methods to the testing dataset was given in Table 1. It shows that the proposed model achieved the second highest average per-class AUC. Note that the model in [15] was trained by using more than 180,000 images from the PLCO dataset [16] as extra training data. Among the methods, which were trained only on the ChestX-ray14 dataset, our ChestNet achieved the most accurate diagnosis of each of those 14 thorax diseases.

**Table 1.** AUC of 4 methods using the official patient-wise split of ChestX-ray14

| Thorax Disease | Wang et al.[6] | Yao et al.[17] | Gundel et al.[15] | ChestNet |
|---|---|---|---|---|
| Atelectasis | 0.7003 | 0.733 | 0.767 | 0.7433 |
| Cardiomegaly | 0.8100 | 0.856 | 0.883 | 0.8748 |
| Effusion | 0.7585 | 0.806 | 0.828 | 0.8114 |
| Infiltration | 0.6614 | 0.673 | 0.709 | 0.6772 |
| Mass | 0.6933 | 0.718 | 0.821 | 0.7833 |
| Nodule | 0.6687 | 0.777 | 0.758 | 0.6975 |
| Pneumonia | 0.6580 | 0.684 | 0.731 | 0.6959 |
| Pneumothorax | 0.7993 | 0.805 | 0.846 | 0.8098 |
| Consolidation | 0.7032 | 0.711 | 0.745 | 0.7256 |
| Edema | 0.8052 | 0.806 | 0.835 | 0.8327 |
| Emphysema | 0.8330 | 0.842 | 0.895 | 0.8222 |
| Fibrosis | 0.7859 | 0.743 | 0.818 | 0.8041 |
| Pleural Thickening | 0.6835 | 0.724 | 0.761 | 0.7513 |
| Hernia | 0.8717 | 0.775 | 0.896 | 0.8996 |
| Average | 0.7451 | 0.761 | **0.807*** | **0.781** |

*The method represented in [15] was trained by using more than 180,000 images from the PLCO dataset [16] as extra training data.

## 5    Discussion

**Contribution of the Attention Branch.** The attention branch allows the proposed ChestNet model to concentrate adaptively on the pathologically abnormal regions, and thus improves its accuracy. If eliminating the attention branch, our ChestNet model would degrade into ResNet-152. Hence, to assess the performance gain caused by incorporating the attention branch into our model, we further compared our model to ResNet-152 with the same experimental settings and listed the per-class AUC, together with the average AUC, of both models in Table 2. It reveals that, with the help of the attention branch, our model overwhelmingly outperformed ResNet-152, improving the average per-class AUC from 0.7401 to 0.7810. Therefore, the attention branch plays a critical role in our model.

**Table 2.** Per-class AUC of the proposed ChestNet model and ResNet-152

| Thorax disease | ResNet-152 | ChestNet |
|---|---|---|
| Atelectasis | 0.6922 | 0.7433 |
| Cardiomegaly | 0.8072 | 0.8748 |
| Effusion | 0.7713 | 0.8114 |
| Infiltration | 0.6401 | 0.6772 |
| Mass | 0.7165 | 0.7833 |
| Nodule | 0.6783 | 0.6975 |
| Pneumonia | 0.6337 | 0.6959 |
| Pneumothorax | 0.7746 | 0.8098 |
| Consolidation | 0.6968 | 0.7256 |
| Edema | 0.8094 | 0.8327 |
| Emphysema | 0.7957 | 0.8222 |
| Fibrosis | 0.7825 | 0.8041 |
| Pleural Thickening | 0.7419 | 0.7513 |
| Hernia | 0.8208 | 0.8996 |
| Average | 0.7401 | **0.7810** |

**Computational Complexity.** It took about 20 hours to train the proposed ChestNet-model and took less than 0.1 second to apply the trained model to diagnose a chest radiograph (Intel Xeon E5-2678V3 ×2, NVIDIA Titan Xp GPU ×4, 128 GB Memory, 120GB SSD and Caffe). It suggests that our ChestNet model, though computationally very complex during the training process that can be performed offline, is very efficient for online testing and could be used in a routine clinical workflow.

## 6    Conclusion

We proposed the ChestNet model, which consists of a classification branch and an attention branch, for the diagnosis of 14 thorax diseases on chest radiographs. We evaluated this model against three state-of-the-art deep learning models on the Chest X-ray 14 dataset using the official patient-wise split. The proposed model achieved an average per-class AUC of 0.781, which is higher than those achieved by other methods, which use no extra training data. Our future work will focus on learning the correlations among those disease image labels and incorporating them into the computer-aided diagnosis process.

## Acknowledge